\documentclass[runningheads]{llncs}
\usepackage{graphicx}
\usepackage{amsmath,amssymb} % define this before the line numbering.
\usepackage{color}
\usepackage{booktabs}
\usepackage{multirow}

\usepackage{epsfig}
\usepackage{subfig}
\usepackage{tabularx}
\usepackage{paralist}
\usepackage{amssymb}

\begin{document}
\pagestyle{headings}
\mainmatter

\title{TopoAL: An Adversarial Learning Approach for Topology-Aware Road Segmentation}

\titlerunning{TopoAL: Adversarial Learning for Topology-Aware Road Segmentation} 
\authorrunning{S. Vasu, M. Kozinski, L. Citraro, and P. Fua} 
\author{Subeesh Vasu, Mateusz Kozinski, Leonardo Citraro, and Pascal Fua}
\institute{CVLab\footnote{This work was funded in part by the Swiss National Science Foundation}, EPFL, Lausanne, Switzerland}

\maketitle

% !TEX root = ../main.tex
% !TEX spellcheck = en-US

\newif\ifdraft
\drafttrue

\definecolor{orange}{rgb}{1,0.5,0}

\newcommand{\bL}[0]{\mathcal{L}}
\newcommand{\bx}[0]{\mathbf{x}}
\newcommand{\bd}[0]{\mathbf{d}}
\newcommand{\by}[0]{\mathbf{y}}
\newcommand{\hby}[0]{\hat{\mathbf{y}}}

\newcommand{\real}{\mathbb{R}}

\newcommand{\PP}{\mathrm{PP}}
\newcommand{\AP}{\mathrm{AP}}
\newcommand{\TP}{\mathrm{TP}}
\newcommand{\IoU}{\mathrm{IoU}}
\newcommand{\SIoU}{\mathrm{SoftIoU}}

\newcommand{\ha}[0]{\hat{a}}
\newcommand{\hb}[0]{\hat{b}}

\newcommand{\comment}[1]{}

%Algos
\newcommand{\RT}{{\it Roadtracer}}
\newcommand{\RTSeg}{{\it Segmentation}}
\newcommand{\SegP}{{\it Seg-Path}}
\newcommand{\DRM}{{\it DeepRoadMapper}}
\newcommand{\RCNN}{{\it RCNN-UNet}}
\newcommand{\RC}{{\it MultiBranch}}
\newcommand{\UNet}{{\it UNet}}
\newcommand{\UNetVGG}{{\it UNet-VGG}}
\newcommand{\LNet}{{\it D-LinkNet}}
\newcommand{\VG}{{\it VanillaGAN}}
\newcommand{\TG}{{\it TopoAL}}
\newcommand{\UNetTG}{{\it UNet-TopoAL}}
\newcommand{\UNetVG}{{\it UNet-VanillaGAN}}
\newcommand{\DRU}{{\it DRU}}
\newcommand{\DRUTG}{{\it DRU-TopoAL}}
\newcommand{\UNetPatchG}{{\it UNet-PatchGAN}}
\newcommand{\PatchG}{{\it PatchGAN}}

\newcommand{\CCQ}{{\it CCQ}}
\newcommand{\TLTS}{{\it TLTS}}
\newcommand{\APLS}{{\it APLS}}
\newcommand{\HM}{{\it H\&M}}
\newcommand{\JUNCT}{{\it JUNCT}}

% !TEX root = ../main.tex
% !TEX spellcheck = en-US

\begin{abstract}

Most state-of-the-art approaches to road extraction from aerial images rely on a CNN trained to label road pixels as foreground and remainder of the image as background. The CNN is usually trained by minimizing pixel-wise losses, which is less than ideal to produce binary masks that preserve the road network's global connectivity. 

To address this issue, we introduce an Adversarial Learning (AL) strategy tailored for our purposes. A naive one would treat the segmentation network as a generator and would feed its output along with ground-truth segmentations to a discriminator. It would then train the generator and discriminator jointly. We will show that this is not enough because it does not capture the fact that most errors are local and need to be treated as such. Instead, we use a more sophisticated discriminator that returns a label pyramid describing what portions of the road network are correct at several different scales. 

This discriminator and the structured labels it returns are what gives our approach its edge and we will show that it outperforms state-of-the-art ones on the challenging RoadTracer dataset. 
 

\end{abstract}
% !TEX root = ../main.tex
% !TEX spellcheck = en-US

\section{Introduction}

Many state-of-the-art algorithms for reconstructing road networks from aerial images approach the problem in terms of foreground/background binary segmentation~\cite{Mnih10,Mnih12,Mattyus17,Cheng17,Batra19}, where the road pixels are the foreground ones. They rely on deep networks and often deliver better performance than approaches that directly predict the road networks as graphs instead of binary masks~\cite{Bastani18,Li18h,Chu19}, even though they fail to account for the connectivity patterns of road networks. There have been several recent efforts at enforcing connectivity constraints on the segmentation outputs by designing topology-aware loss functions~\cite{Mattyus18,Mosinska18} or by relying on multi-task learning~\cite{Batra19,Yang19}. These approaches to enforcing connectivity on the output of a binary segmentation algorithm are mostly implicit: The network or the loss functions are modified in such a way that the resulting segmentations yield a more road-like connectivity.

% !TEX root = ../main.tex
% !TEX spellcheck = en-US

\begin{figure}
\centering
\includegraphics[height=5cm]{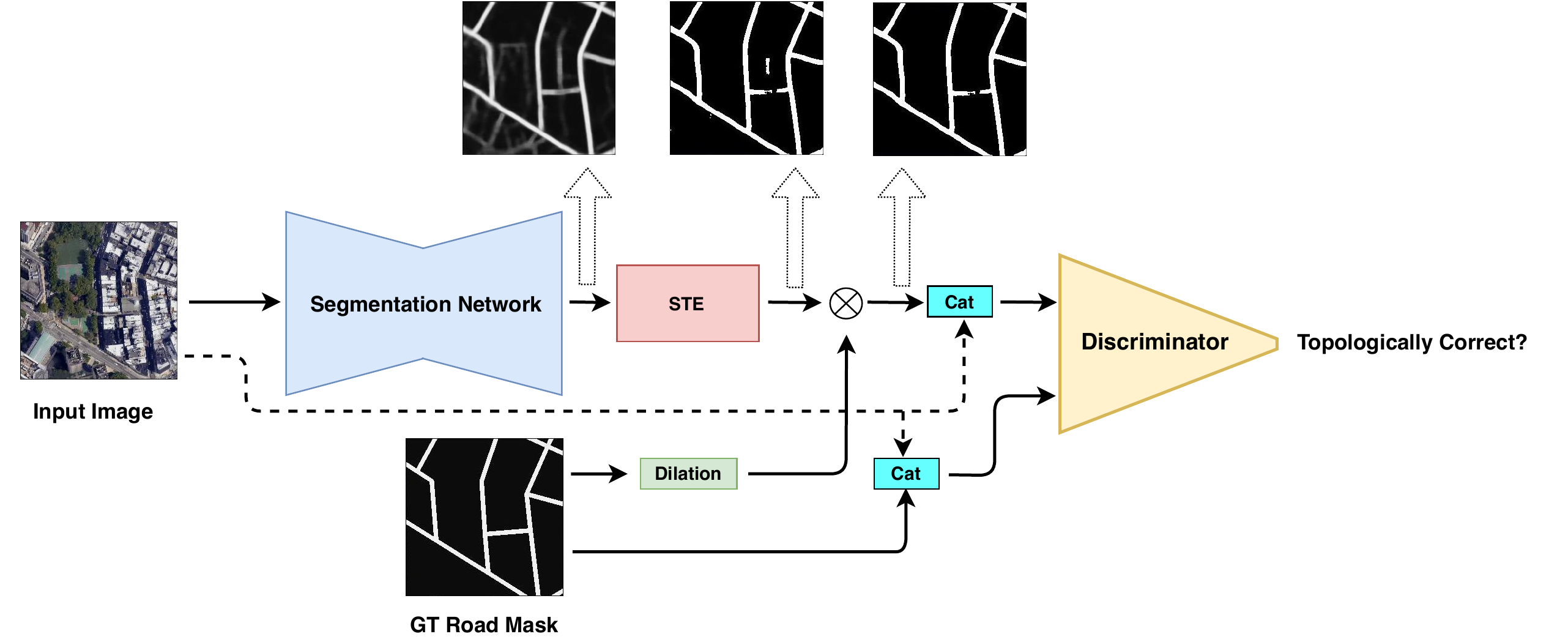}
\caption{{\bf Network Architecture.} We use a segmentation network to predict the road probability maps which is then passed through a straight through estimator (STE) to generate binarised predictions. This is followed by multiplication with dilated ground truth masks to generate prediction based input to the discriminator. Another input to the discriminator is the ground truth mask which is used only during discriminator training. The road masks are concatenated with the input image before feeding them to the discriminator. Discriminator is then trained to predict spatial-aware dynamic decisions on the topological correctness of inputs.}
\label{fig:topogan_net_arch}
\end{figure}

\begin{figure*}[t]
  \centering
  \includegraphics[scale=0.31, trim={1.5cm 0cm 1.5cm 0cm}, clip]{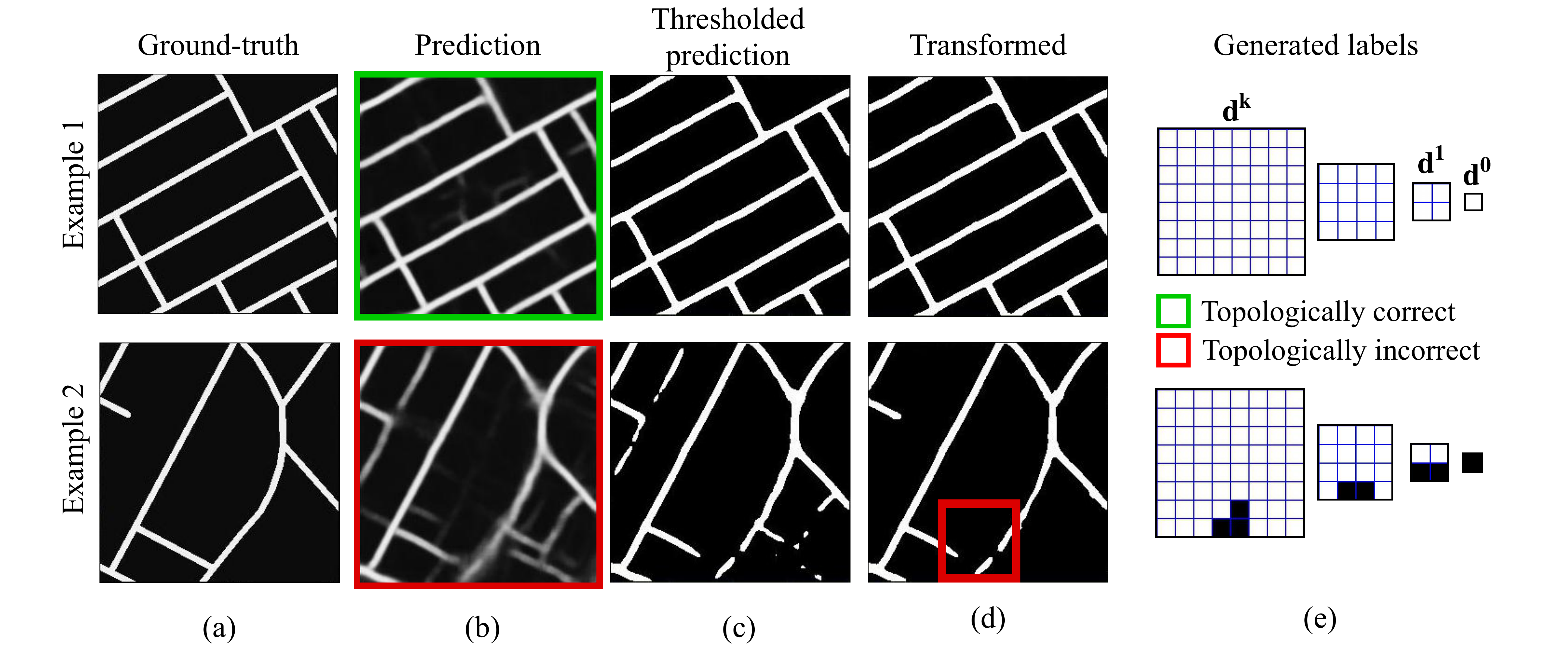} 
  \caption{{\bf Label generation for discriminator training.} Two example patches with the corresponding multi-scale labels. One patch is topologically correct, the other contains an interruption. The labels are generated from the processing of ground-truth and predicted masks.  (\textbf{a}) Ground truth road masks and corresponding (\textbf{b}) predictions from segmentation network. (\textbf{c}) Thresholded predicted masks. (\textbf{d}) Masks generated by multiplying (c) with the dilated form of (a). (\textbf{e}) Spatial-aware labels made by comparing (a) with (d). White and black boxes denote label 1 (correct topology) and 0 (topology error).}
  \label{fig:dynamic_labels}
\end{figure*}

In this paper, we propose a different and more explicit approach that relies on Adversarial Learning (AL). We use the training methodology of Generative Adversarial Networks (GAN) depicted by Fig.~\ref{fig:topogan_net_arch} to reduce topological discrepancies between the probability maps produced by our segmentation network and that of real road networks. A naive way to do this would be to treat the segmentation network as a generator and to feed its output along with ground-truth segmentations to a discriminator and then to train them jointly. Unfortunately, we will show that this approach is too global for the discriminator to learn to detect local topological errors and for the segmentation network to avoid making them. To remedy this, we introduce the following two key modifications:
\begin{itemize}

 \item {\bf Spatially aware labels.} Labeling a delineation as globally correct or incorrect is too coarse. As shown in Fig.~\ref{fig:dynamic_labels}, the discriminator returns a label pyramid that describes what portions of the road network are correct at several different scales. 
 
 \item  {\bf Dynamically assigned labels.} These correctness decisions are not made {\it a priori}. Instead, the segmentations produced by the generator are evaluated for correctness in the course of the training procedure. 

\end{itemize} 
Our main contribution is therefore a novel AL strategy for imroving connectivity constraints on the output of road segmentation networks. We will refer to it as \TG{}. We will use the RoadTracer dataset~\cite{Bastani18} to show that it compares favorably to state-of-the-art methods.

% !TEX root = ../main.tex
% !TEX spellcheck = en-US

% !TEX root = ../main.tex
% !TEX spellcheck = en-US

\section{Related Work}
\label{sec:related}

Most of the early approaches to extracting road networks relied on handcrafted features and prior knowledge about road geometry to optimize complex objective functions~\cite{Barzohar96,laptev00,stoica04,Wegner13,Chai13,Turetken13a}, to quote only the most recent ones. They have now been mostly superseded by deep learning techniques. One of the first such method is the approach of~\cite{Mnih10} in which image patches were fed as input to a fully connected neural network. Due to memory constraints, only limited context information could be exploited. The advent of convolutional neural networks (CNNs) opened the door to increasing the size of the receptive fields. 

Many state-of-the-art approaches formulate road extraction as a segmentation problem and rely on encoder-decoder architectures such as U-Net~\cite{Ronneberger15}, D-LinkNet~\cite{Zhou18b}, or recurrent versions of these architectures~\cite{Yang19,Wang19c}. While these approaches feature large receptive fields, none of them explicitly takes into account the connectivity of the resulting binary masks. Several recent approaches have attempted to remedy this. A topology-aware loss function was introduced in~\cite{Mosinska18,Mattyus18} while the use of an additional centerline extraction module was proposed in~\cite{Yang19}. In~\cite{Batra19}, orientation prediction is used as an auxiliary task to improve the connectivity of the predicted masks.~\cite{Mattyus17} introduced a post-processing algorithm to reconstruct the graph from segmentation outputs by reasoning about missing connections and applying a number of heuristics. A unified approach to segmentation and connectivity reasoning was presented in~\cite{Mosinska19}. It uses a segmentation network and a classifier that shares the same encoder representation. The classifier is used to reason out the connectivity in the segmentation output. Finally,~\cite{Li18h} proposed an approach to predict road segments in the form of vector representation instead of pixel-wise segmentations. Their approach use a CNN to extract key-points and edge evidences from a given patch, which were then fed sequentially to a recurrent neural network (RNN) to produce vector representations of the underlying road segments.

A radically different approach is to directly build the graph without segmenting. This is typically done iteratively by adding road segments one by one. In~\cite{Ventura18}, a CNN is trained to predict the local connectivity among the central pixel of an input patch and its border points. To reconstruct the road network corresponding to the whole image, the algorithm iteratively performs patch-wise identification of input and exit points and the associated connections. In~\cite{Bastani18}, a CNN-based decision function is used to guide an iterative search process, which starts from a search point known to be part of the road network. At each step, the CNN takes the point and the already reconstructed roads in its neighborhood as input and outputs the decision to walk a fixed distance along a certain direction or to stop and return to previous search point. In a similar spirit, a Neural Turtle Graphics can be used to iteratively generate new nodes and the corresponding edges connecting to the existing nodes~\cite{Chu19}. This approach relies on RNNs instead of CNNs, as in~\cite{Ventura18,Bastani18}.

Both segmentation based approaches and graph-based approaches have some clear vulnerabilities. The former are subject to small-scale topological errors in the form of missing connections. The latter, though free from topological errors, are vulnerable to error propagation due to the iterative reconstruction policy. We will show that our proposed approach is less affected by these difficulties as compared to other state-of-the-art algorithms.

Some of the recents works have proposed to use multiple discriminators in generative adversarial networks setups~\cite{Chen18d,Iizuka17,Wang18c}. A domain adaptation technique for semantic segmentation is proposed in \cite{Chen18d}. They divide the discriminator input into different spatial regions, and associate different classifiers to each region. The concepts of multi-scale discriminator~\cite{Wang18c} and local-global discriminators~\cite{Iizuka17} were introduced to examine the input data at different context levels. Unlike all these methods, our approach differs in multiple aspects including label generation, the use of a single network architecture, and the input characteristics that the discriminator has to learn.
% !TEX root = ../main.tex
% !TEX spellcheck = en-US

\section{Method}

As shown in Fig.~\ref{fig:topogan_net_arch}, our approach closely traces the structure of a GAN~\cite{Goodfellow14b}. For this reason and simplicity in the terminology, we refer to the segmentation network as the generator and to the evaluator of the delineated predictions as the discriminator. In a traditional AL, the discriminator would assign a binary label to the segmentations it produces. However, this is not suitable for our purpose because the predicted mask can be correct almost everywhere except for few locations that result in poor connectivity. To properly account for this and to provide spatially-aware supervision, our discriminator predicts the more sophisticated labels depicted by Fig.~\ref{fig:dynamic_labels}. They are computed online by splitting the image into increasingly fine partitions and then labeling each element of these partitions as valid or not given the ground-truth data. The labels corresponding to ground truth masks have the same structure but are uniformly valid and the corresponding loss function takes all labels at all scales into account.

The combination of using these more sophisticated labels and computing them dynamically during training, instead of fixing them {\it a priori} as is usually done is what gives our approach its edge. We now formalize it and describe its individual components in turn. 

\subsection{Formalization}

Let $\bx \in \real^{H \cdot W \cdot C}$ be a $C$-channel input image of size $H\times W$, and let $\by \in\{0, 1\}^{H\cdot W}$~be the corresponding ground-truth road mask, with $1$~indicating pixels corresponding to a road and $0$~indicating the background pixels. Let us consider a segmentation network $G$~that takes $\bx$~as the input and outputs a {\it probability map} $\hby = G(\bx) \in [0, 1]^{H\cdot W}$. For any given pixel~$i$, $\hby_i$ is taken to be the probability that it is a road pixel. The weights of the network are typically learned by minimizing the pixel-wise binary cross-entropy (BCE) loss
\begin{equation}
\bL_{bce}(\hby, \by) = -\sum_{i} \left[(1-\by_i)\cdot\log (1-\hby_i)\right. \left.+ ~\by_i\cdot\log \hby_i\right].
\label{eq:bceLoss}
\end{equation}
Such a loss function penalizes mistakes everywhere equally regardless of their influence on the underlying geometry of the predicted road network. To remedy this, we use the segmentation network $G$ as the generator of the GAN of Fig.~\ref{fig:topogan_net_arch} and define a discriminator network $D$ whose role is to identify topological errors in the generator output. $G$ and $D$ are trained by making them play a game: The weights of $G$ are optimized to generate segmentations that $D$ cannot distinguish from ground-truth ones and the weights of $D$ are optimized to make that distinction as well as possible.

\subsection{Discriminator Network: $D$}
\label{secn:discriminator}

In a traditional GAN, $D$ is trained to return a binary label that is 1 if a sample is a ground-truth one and 0 if it is one generated by the $G$ network. We will show in the result section that this does not help much in our case, mostly because the errors that $G$ makes are local and that a single binary label is not enough to characterize them. Instead, we designed $D$ to classify different portions of the predicted masks at different scales as correct or not. This pyramid approach allows the discriminator to provide both local and global supervision to the generator and to model spatial dependencies between neighboring locations. 
We now describe the label generation and the discriminator architecture in detail.

\subsubsection{Scale Space Labels.} The most challenging aspect  of road network reconstruction is recovering the connectivity of the network by avoiding topological errors. They most often manifest themselves as short breaks in road segments that spoil an otherwise mostly correct binary mask. An effective discriminator must therefore detect and localize such mistakes so that the generator can fix them. 

To this end, we define spatially-aware labels as follows. We consider a pyramid of increasingly zoomed-out versions of the predicted mask, as shown in Fig.~\ref{fig:dynamic_labels}(e).  At level $k$ of the resulting pyramid, we divide the mask into non-overlapping patches of size $H_k\times W_k$ and associate a binary value to  each one, depending on the topological correctness within it. The label matrix generated at level $k$ is of size $\frac{H}{H_k}\times \frac{W}{W_k}$. The collection of such label matrices can be regarded as a representation of topological correctness at different scales and locations.  

In practice, our input images are of size $256\times256$ and we have used four levels with $H_k = W_k = \{256, 128, 64, 32\}$. At one extreme of the range, we assign a single label to the whole image and, at the other, we assess the correctness within $32\times32$ patches. As shown in Fig.~\ref{fig:dynamic_labels}(e), for a road mask that has topological error, the values in the label matrices are neither all zeros nor all ones. Instead, they encode fine- or coarse-level locations of topological errors in the generators output. This allows the discriminator to effectively use multi-scale information.

\subsubsection{Dynamic Label Assignment.}

There is no way to know {\it a priori} in which of the patches discussed above the topology is correct. Therefore, unlike in traditional GANs, we cannot predefine the labels we assign to the generator output. Instead, we must do this dynamically for each prediction made by the generator. 

For example, the patches outlined in green or red in Fig.~\ref{fig:dynamic_labels} are deemed correct or incorrect, respectively. To make this assessment, we compare the probability map produced by the generator to the ground-truth within the patch of interest. Alongside, to match with the labelling strategy, we use a transformed form of $\hby$ to construct the corresponding input to the discriminator. To generate the inputs and labels for the discriminator training we use the following steps:
\begin{enumerate}
 
  \item {\it Differentiable Thresholding.} We binarize the probability map produced by the generator. To preserve differentiability, we use STE~\cite{Bengio13b,Yin19b} that thresholds during the forward pass while behaving as the identity function during the backward pass. We set the threshold to be 0.5 in our experiments.
    
  \item {\it Multiplication by the dilated ground truth.} The generator can produce false negatives---road pixels that are classified as background as highlighted in Fig.~\ref{fig:dynamic_labels}(d)---and false positives---background  pixels that are classified as road pixels as highlighted in Fig.~\ref{fig:dynamic_labels}(c). The false negatives are those that cause disconnections and break the connectivity of the network. Furthermore, it is not unusual for some real roads to be missing from the ground-truth. We have therefore found empirically that ignoring the false positives and focusing on the false negatives to be beneficial. Before feeding the thresholded prediction to the input of the discriminator, we  therefore  multiply it with a dilated version of the ground truth mask. The dilation accounts for the fact that the centerline locations are not always precise in the ground truth. We set the dilation factor to 3.  Let us denote the resulting mask as $T_0(\hby)$. Two examples are shown in Fig.~\ref{fig:dynamic_labels}(d).
  
  \item{\it Concatenating with Input Image.} This final step is used to generate the complete discriminator input. The ground-truth road masks may contain genuine interruptions, for example because of road dead ends near to other road sections as in the top-left corner of second row of Fig. 2 (d). To help the discriminator distinguish these from unwarranted ones (i.e., erroneous interruptions of the predicted road network as in the red box in Fig. 2 (d)), we also feed it the input image $\bx$ so that it can examine the context in which these interruptions occur. To this end, we concatenate $\bx$ with the road mask before feeding it to the discriminator.  We will refer to the discriminator formed by concatenating $T_0(\hby)$ with  $\bx$ as $T(\hby)$.
 
   \item{\it Assigning Label Values.} To generate the label values, we compare the ground truth skeleton with the prediction $T_0(\hby)$ and count the false nagatives (number of pixels in the skeleton that are not covered in $T_0(\hby)$). We use the count of false negatives to identify the cell that is likely to contain topological errors. We assign zero to these patches and a 1 otherwise.

 \end{enumerate}
 The operations corresponding to STE, dilation, and concatenation are represented by the pink, green, and blue boxes in Fig.~\ref{fig:topogan_net_arch}. Detailed illustration on the proposed label generation scheme can be found in Fig.~\ref{fig:dynamic_labels}.

\subsubsection{Architecture.}

% !TEX root = ../main.tex
% !TEX spellcheck = en-US

\begin{figure*}[t]
  \centering
\begin{tabular}{c}
  \includegraphics[width=0.37\textwidth, angle=270, trim={4.3cm, 0cm, 4.3cm, 0cm},clip]{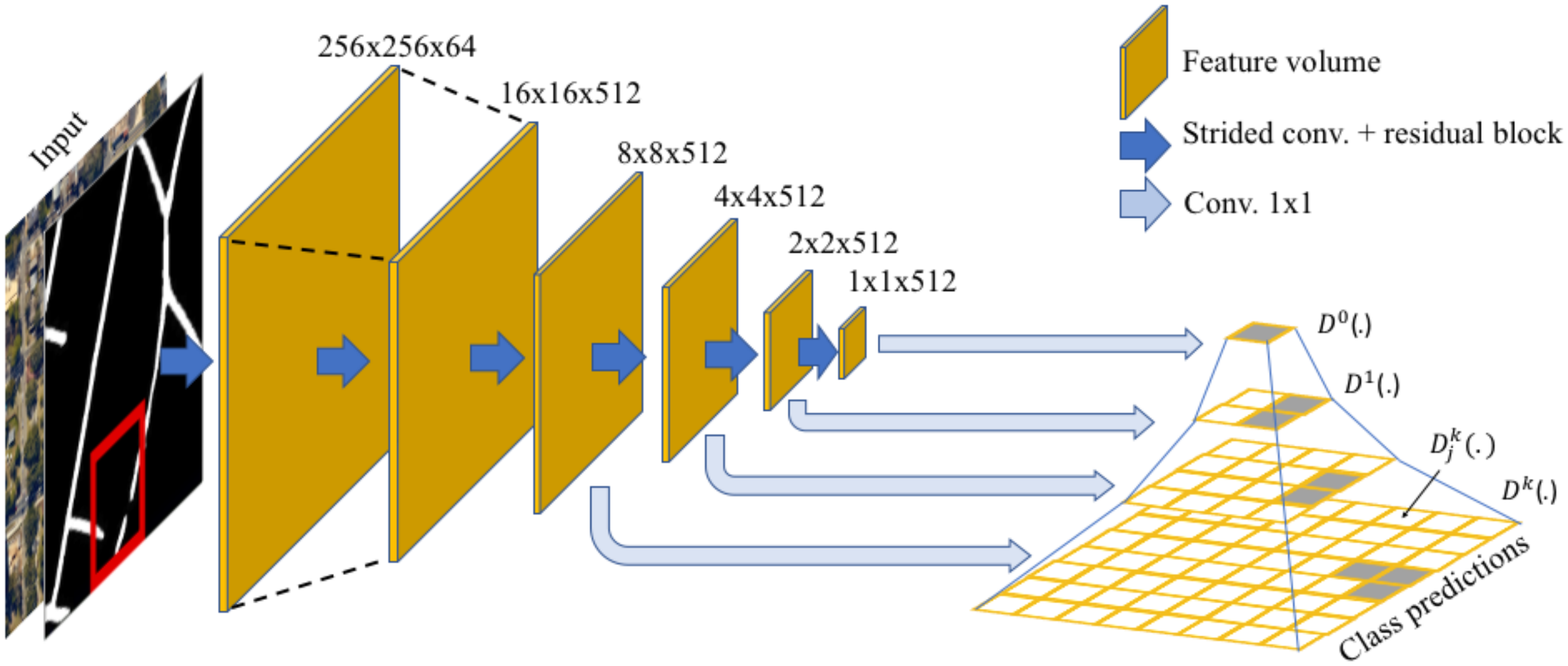} 
\end{tabular}
  \caption{{\bf Discriminator.} {\small The discriminator is a single fully convolutional network that downsample the input to the desired scales and locations. Each downsampling stage is composed of a convolutional layer with stride factor 2, followed by a residual block. The first convolution layer is set to have 64 feature channels. The channel number doubles after every strided convolution until it reaches 512. The class prediction are produced from the latest stages using a $1 \times 1$ convolutional layer.}}
  \label{fig:discriminator}
\end{figure*}

To implement $D$, we use a fully convolutional network similar to PatchGAN~\cite{Isola17}, but with four outputs, each having a different resolution. As a result, $D$ outputs  a pyramid of probability maps having the same structure as the spatially aware labels. Fig.\ref{fig:discriminator} depicts its architecture. It comprises eight stages that downsample to $\frac{1}{256}^{th}$ of the input resolution. Each stage is made of a convolutional layer with stride factor 2 followed by a residual block. The first convolution layer has 64 feature channels. The channel number doubles after every strided convolution until stage 4; the number of channels are kept at 512 afterward. The final residual block is followed by a single convolutional layer to output the prediction at the last stage. Features from the output of residual blocks at stages 5-7 are passed through a single convolutional layer to generate predictions at respective resolutions.

\subsection{Generator Network: $G$}
\label{secn:generator}

One could use any standard segmentation network as the generator $G$. To demonstrate this, we experimented with two different ones, a standard \UNet~\cite{Ronneberger15} and a recurrent version~\cite{Wang19c} of it, which we will refer to as \DRU. 

\UNet{} is a fully convolutional encoder-decoder network with skip connections that serves as the backbone of several recent road extraction algorithms~\cite{Mosinska18,Mosinska19}. We follow the standard \UNet{} design with four levels. Each one comprises two convolutional operations followed by max pooling. We set the first convolutional unit to have 64 feature channels. \DRU~\cite{Wang19c} is a \UNet{} with Dual-gated Recurrent Units. 
It performs recursion on the input-output and in multiple internal states of the network to improve the overall performance with only minimal increments in model size. In the original paper, a lightweight \UNet{} architecture with 32 channels was used. Here we use 64 channels as in the standard \UNet{}.

By combining our proposed discriminator with the two generator networks \UNet~and \DRU, we build two different methods and dub them as \UNetTG{} and \DRUTG{} respectively.

\subsection{Training}
\label{secn:training}

To train the generator $G$ and discriminator $D$, we follow the usual GAN approach, which is to alternatively minimize the loss functions with respect to the generator and the discriminator. The difference, however, resides in the discriminator that takes as input $T(\hby)$, the generator's output transformed as described before, and $c [ \by, \bx ]$ formed by concatenating the ground truth mask $\by$ with $\bx$. We train the discriminator to minimize the BCE loss between $T(\hby)$ outputs and the corresponding spatially aware labels. We write the loss function as 
\begin{equation}
\bL_{D1}(\hby, \by) =  \sum_{k} \bL_{bce}(D^k(T(\hby)), \bd^k) + \sum_{j} -\log D^k_j(c [ \by, \bx ]) \; ,
\label{eq:TopoganDiscLoss}
\end{equation}
where $k \in\{0, 1, 2, 3\}$ is used to index the four outputs from the discriminator.  $D^k_j$ refers to $j^{th}$ element of $k^{th}$ discriminator output, and $\bd^k$ is the spatial-aware labels for the level $k$.

The generator is trained using a combination of BCE Loss and adversarial loss given by
\begin{equation}
\bL_{G1}(\hby, \by) = \bL_{bce}(\hby, \by) +  \lambda_{A}\sum_{k} \sum_{j} -\log D^k_j(T(\hby)) \; ,
\end{equation}
where $\lambda_{A}$ is a scalar weight that controls the influence of adversarial loss. In our experiments we set $\lambda_{A}$ to be 0.005, an empirically found optimal value.

Note that if we use a  single discriminator output at level k = 0 in Eq.~\ref{eq:TopoganDiscLoss}, predefined label value of 0 for the generator output, that is, $d^0$ = 0, and ($\hby$, $\by$) as the two inputs to the discriminator network, this reduces to a standard GAN. We will use this as a baseline that we will refer to as \VG{}.
% !TEX root = ../main.tex
% !TEX spellcheck = en-US

\section{Experiments}

We now describe the dataset we have tested our approach on, the performance measures we used to assess the quality of the reconstructions, and baselines we used for comparison purposes. We then show that our approach can be used to enhance the performance of two of them and report our performance against that of the others. Additional experimental results including performance comparison on the DeepGlobe Dataset~\cite{DeepGlobe18} and the ablation studies revealing the importance of each component of our approach are provided in the supplementary material.

\subsection{Datasets, Metrics, and Baselines}

\paragraph{Dataset.} We perform our experiments on the RoadTracer dataset~\cite{Bastani18}, which is one of the most recently published, largest, and most challenging road network dataset. It contains high-resolution satellite images covering the urban areas of forty cities in six different countries. The labels are generated using OpenStreetMap in the form of graphs. It covers areas featuring highways, urban roads, and rural paths, which results in extreme appearance variations. The roads are often occluded by trees, buildings, and shadows, making it difficult for segmentation approaches to preserve topology. Finally, the set of 25 training cities and 15 testing cities are totally disjoint, which makes generalization more difficult than if the training and testing images were from the same city. 

\paragraph{Metrics.} Many metrics have been developed to compare the estimated road networks to that of the ground truth. These metrics can be broadly classified into two categories, pixel-based and topology-based. We use both kinds for the sake of completeness. 
\begin{itemize}

\item \textbf{Correctness/Completeness/Quality (\CCQ{})}. This is a pixel-based metrics intended to measure the similarity between a skeletonized prediction and the corresponding annotation. In segmentation tasks, predictions are often evaluated using precision $=\frac{| \TP | } {| \PP |}$, recall $=\frac{|\TP|}{| \AP |}$, and intersection-over-union $=\frac{\TP}{\PP\cup\AP}$, where  $\PP$ is the set of foreground pixels in the prediction, $\AP$ is the set of foreground pixels in the ground truth, and $\TP  = \PP \cap \AP$. To account for the shift between predictions and the ground truth, precision, recall, and intersection-over-union have been relaxed~\cite{wiedemann98}. The resulting quantities are named correctness, completeness, and quality respectively. To assess the performance using a single metric we report the values of quality (qual.). In our experiments, we set the allowable pixel shift to 2 pixels.

\item \textbf{Too Long/Too Short (\TLTS{}}). \TLTS{}~\cite{Wegner13} compares the lengths of the shortest paths between randomly sampled ground-truth nodes matched to the prediction. If the length of the path in the predicted graph is within 5$\%$ of that of the path in the ground-truth the path is declared to be correct. We use the percentage of correct paths, denoted as \TLTS{}-corr, to assess the prediction quality. We set the threshold defining if nodes from the ground-truth are matched to the prediction to 25.

\item \textbf{Average Path Length Similarity (\APLS{}}). \APLS{}~\cite{VanEtten18} aggregates the differences in optimal path lengths between nodes in the ground truth and predicted graphs. The Average Path Length Similarity is computed as 
\begin{equation}
1-\frac{1}{N}\sum \min\left\{1, \frac{L(a,b)-L(\ha,\hb)}{L(a,b)}\right\}
\end{equation}
where $a$ and $b$ are nodes in the ground truth graph, $\ha$ and $\hb$ are the corresponding nodes in the predicted graph, $N$ is the number of nodes in ground truth, and $L$ denotes the length of the shortest path connecting them. To penalize false positives, the same procedure is repeated by swapping ground-truth and prediction. The final score is the harmonic mean of the two.

\item \textbf{Junction (\JUNCT{}}). \JUNCT{}~\cite{Bastani18} compares the degree of corresponding nodes with at least three incident edges, called junctions. The correspondences are established greedily by matching closest nodes. For each ground-truth junction $v$ that is
matched to a predicted junction $u$, the per-junction recall $f_{u,correct}$ is computed by taking into account the fraction of edges incident on $v$ that are also captured around $u$. The same operation is performed to compute the per-junction 1-precision $f_{v,error}$ which is the
fraction of edges incident on $u$ that do not appear around $v$. For this metric we report the f1-score compute using $f_{u,correct}$ as recall and $1-f_{v,error}$ as precision. We used the implementation provided in~\cite{Bastani18} with default parameters. 

\item \textbf{Holes and Marbles (\HM{}}). This metric first extracts small subgraphs from the ground-truth and match them to the prediction. Then, compares sets of locations in the two subgraphs accessible by traveling a predefined distance away from a randomly sampled point. Virtual control points, namely holes, are dropped at regular intervals along the paths in the ground truth graph. The same process is repeated in the predicted graph, these control points are called marbles. A hole is said to be matched if it lies sufficiently close to one of the marbles. The process is repeated for many subgraphs and the total count of matched and unmatched points are then used to compute precision and recall. To asses the prediction quality using a single value, we report the f1-score as in~\cite{Biagioni12}.
We set the threshold defining if nodes from the ground-truth are matched to the prediction to 25. We extracted subgraph of radius 300 and sampled 1000 of them for each sample.

\end{itemize}

\paragraph{Baselines and Variants.} 

We use the following approaches that are briefly described in Section~\ref{sec:related} as baselines.
\begin{itemize}
\setlength\itemsep{1mm}
\item \UNet~\cite{Ronneberger15}:    Fully-convolutional network with skip connections.
\item \UNetVGG~\cite{Mosinska18}:  {Fully-convolutional network with skip connections and \\ topology-aware loss function.}
\item \DRU~\cite{Wang19c}:               Fully-convolutional network with skip connections and recursion. 
\item \SegP~\cite{Mosinska19}:         Two-branch network that jointly learns to segment linear structures and to classify candidate connections.
\item \RC~\cite{Batra19}:                   Recursive architecture jointly trained for road segmentation and orientation estimation.
\item \RCNN~\cite{Yang19}:               Fully convolutional network with recursive convolutional layers.
\item \DRM~\cite{Mattyus17}:           A fully convolutional network based segmentation followed by heuristics based post-processing to generate the graph.
\item \RT~\cite{Bastani18}:                Graph constructed iteratively with new node locations being selected by a convolutional network.
\end{itemize}
We compare these baselines against three variants of our approach introduced in Sections~\ref{secn:generator} and \ref{secn:training}.
\begin{itemize}
\setlength\itemsep{1mm}
 \item \DRUTG: We use the network of \DRU~\cite{Wang19c} as our generator.
 
 \item \UNetTG: We use the network of \UNet~\cite{Ronneberger15} as our generator. 
 
 \item \UNetVG: We use the network of \UNet~\cite{Ronneberger15} as our generator and replace our sophisticated discriminator by a simple one that returns a simple binary flag for each input mask. 

\end{itemize}

\subsection{Implementation details}

To train our \TG{} networks we rendered the RoadTracer ground-truth graphs to half resolution to generate pixel-wise annotations. We have experimented with different input sizes and observed that half-resolution produced the best results on the Roadtracer dataset when the network is trained using binary cross-entropy loss (BCE) alone. We take this to mean that the higher resolution details of this dataset are not key to producing globally correct topologies, and the half-resolution provides the optimal trade-off between the required details in the input image and the effective context available to the network. As input, we used $256\times256$ patches randomly cropped from the training images. To improve the generalization of learned network, we employed data augmentations in the form of random horizontal flip, vertical flip, scaling and rotation. To train both the generator and discriminator, we used the Adam optimizer~\cite{Kingma14a} with a $10^{-4}$  learning rate and a batch size of 4. All the models are implemented in Pytorch~\cite{Paszke17}. To construct the dynamic labels, a 32x32 patch is declared topologically incorrect if it contains an erroneous road interruption of at least 4 pixels long. Larger patches are declared incorrect if they contain an incorrect 32x32 sub-patch. We selected this threshold based on visual inspection to separate road interruptions from misalignment of the predicted and annotated roads.

We trained \UNet{}, \UNetVGG{} and \DRU{} using the same settings as \TG{}, as described in Section \ref{secn:generator}. For \DRU{} and \DRUTG{}, we set the number of recursions to 3 and use the sum of losses corresponding to outputs from all the recursions. For \DRUTG{}, outputs from all the recursions is used for the discriminator training, and the total adversarial loss is computed as the sum of losses corresponding to all the recursions. We retrained the \RC{} network on the RoadTracer dataset using the code provided by the authors with default settings. For \DRM{}, we use the results shared by the authors of \cite{Bastani18}. For all other methods we use the predicted road networks made publicly available by the authors.

% !TEX root = ../main.tex
% !TEX spellcheck = en-US

\begin{table*}[t]
\centering
\resizebox{0.88\linewidth}{!}{%
	\setlength{\tabcolsep}{3pt}
	% junction corr comp 2l2s apls hm junction v2 2l2s v2
	\begin{tabular}{@{}  l l p{0.2cm}  ccc c c c c c c@{} }
	%\toprule
	
\cmidrule{4-12} 
	
 & & &
\multicolumn{1}{c}{\small Pixel-based} &&
\multicolumn{7}{c}{\small Topology-aware}\\

\cmidrule{4-4}
\cmidrule{6-12}  

 & & &
\multicolumn{1}{c}{\small CCQ} &&
\multicolumn{1}{c}{\small TLTS} &&
\multicolumn{1}{c}{\small APLS} &&
\multicolumn{1}{c}{\small JUNCT} &&
\multicolumn{1}{c}{\small \HM{}}\\

& Method & &
qual. &&
corr. &&
&&
f1 &&
f1\\
%\midrule
\cmidrule{1-2} 	
\cmidrule{4-12} 	

\multirow{6}{*}{\rotatebox{90}{RoadTracer}}

&      \UNet{}~\cite{Ronneberger15} & &
        0.632 &&  % CCQ
        0.323 && % TLTS
        0.619 && % APLS
        0.792 &&  % JUNCT
        0.737 % SUBG
        \\
 &
        \UNetVG{} & &
        0.636 &&  % CCQ
        0.328 && % TLTS
        0.607 && % APLS
        0.776 &&  % JUNCT
        0.748 % SUBG
        \\

&
        \UNetTG{} (Ours) & &
        \bf{0.658} &&  % CCQ
        \bf{0.388} && % TLTS
        \bf{0.666} && % APLS
        \bf{0.808} &&  % JUNCT
        \bf{0.767} % SUBG
        \\ 
\cmidrule{2-12} 
&
        \DRU~\cite{Wang19c} & &
        0.656 &&  % CCQ
        0.437 && % TLTS
        0.697 && % APLS
        0.821 &&  % JUNCT
        0.768 % SUBG
        \\

&
        \DRUTG{} (Ours) & &
        \bf{0.657} & &  % CCQ
        \bf{0.480} && % TLTS
        \bf{0.725} && % APLS
         \bf{0.837} &&  % JUNCT
        \bf{0.787} % SUBG
        \\

\cmidrule{1-12}

\end{tabular}
}
\caption{
Quantitative comparison between baselines segmentation networks, our improved versions, and \UNetVG{}. Our \TG{} approach improves the baselines on all metrics. On the other hand, \UNetVG{} performance is only comparable to that of the baseline network \UNet{}.
\label{tab:tabel_comp}
}
\end{table*}

\subsection{Boosting the Performance of Existing Architectures}

In Table~\ref{tab:tabel_comp}, we compare \UNet{} and \DRU{} against \UNetTG{} and \DRUTG{}, which use  \UNet{} and \DRU{}  as their generators. In both cases, we can see our scheme consistently boosts their performance, especially the  \TLTS{}, \APLS{}, \JUNCT{} and \HM{} metrics that are designed to assess topological correctness. The first row of Fig.~\ref{fig:qual_comparison}  provides corresponding qualitative results. We also report the performance of \UNetVG{}, which implements a standard GAN, in Table~\ref{tab:tabel_comp}. It can be noted that it is not better than its generator \UNet{}. Therefore, using a GAN by itself does not bring the same improvements as \TG{}. As is evident from the first row of Fig.~\ref{fig:qual_comparison}, adding our topology loss not only refine the predictions but also improves the topological correctness. On the other hand, predictions from \UNetVG{} does not respect the topological correctness.

\subsection{Comparison against the State-of-the-Art} 

We now turn to compare our results to those of all the baselines discussed above and report the results in Table~\ref{tab:tabel_comp_all}. We provide corresponding qualitative results in the last three rows of 
Fig.~\ref{fig:qual_comparison}.

\DRUTG{} does best on three of the four topology-aware metrics and is second on the fourth, without any of the post-processing that \SegP{} perform. For the pixel-based measures, \DRUTG{} performs honorably but does not truly dominate the other algorithms. This is not surprising because \TG{} targets small interruptions in road segment. They are few in numbers but critical in terms of topological correctness. Since CCQ only performs pixel-wise comparisons, it is relatively insensitive to the kind of errors we detect and fix.

As indicated by the example in the second row of Fig.~\ref{fig:qual_comparison}, adding VGG loss (UNet-VGG) helps to suppress spurious prediction errors but does not lead to significant improvement in the prediction of true roads. From the third row of Fig.~\ref{fig:qual_comparison} one can observe that, despite the multi-tasking strategy, \RC~fails to predict the roads at occluded regions (highlighted in blue), a limitation that was reported in~\cite{Batra19}. On the other hand, \UNetTG~and \DRUTG~depicts progressive improvements towards filling up such gaps. In comparison with the proposed methods, \RT~and \SegP~that use trained networks to enforce connectivity explicitly, either fails to predict some road segments completely (highlighted in blue) or result in big false positives (highlighted in green) that maintains connectivity to the other parts of the predicted network. The last row of Fig.~\ref{fig:qual_comparison} is an exceptional case wherein trees mostly occlude true roads.\footnote{Ground truth mask does not show most of the actual roads because of the omission noise.} As is evident, among the segmentation methods, \DRUTG~shows the most promising result against occlusion effects, while maintaining favorable performance against~\SegP.

% !TEX root = ../main.tex
% !TEX spellcheck = en-US

\begin{table*}[t]
\centering
\resizebox{0.88\linewidth}{!}{%
	\setlength{\tabcolsep}{3pt}
	% junction corr comp 2l2s apls hm junction v2 2l2s v2
	\begin{tabular}{@{}  l l p{0.2cm}  ccc c c c c c c@{} }
	%\toprule

\cmidrule{4-12} 
	
 & & &
\multicolumn{1}{c}{\small Pixel-based} &&
\multicolumn{7}{c}{\small Topology-aware}\\

\cmidrule{4-4}
\cmidrule{6-12} 

 & & &
\multicolumn{1}{c}{\small CCQ} &&
\multicolumn{1}{c}{\small TLTS} &&
\multicolumn{1}{c}{\small APLS} &&
\multicolumn{1}{c}{\small JUNCT} &&
\multicolumn{1}{c}{\small \HM{}}\\

& Method & &
qual. &&
corr. &&
&&
f1 &&
f1\\
%\midrule
\cmidrule{1-2} 	
\cmidrule{4-12} 	

\multirow{10}{*}{\rotatebox{90}{RoadTracer}}

&      \DRM~\cite{Mattyus17} & &
        0.435 &&  % CCQ
        0.069 && % TLTS
        0.247 && % APLS
        0.514 &&  % JUNCT
        0.469 % SUBG
        \\

&       \RT~\cite{Bastani18} &&
        0.477 &&
        0.420 &&
         0.591&& 0.812&& 0.714\\
 
&        \RCNN~\cite{Yang19}  &&  0.628 && 0.201 && 0.474 &&  0.790 && 0.701  \\

&	    \RC~\cite{Batra19}     &&  \bf{0.659} && 0.439 && 0.682 && 0.798 && 0.765 \\

&        \SegP~\cite{Mosinska19}  &&  0.535 && \bf{0.489} && 0.679 && 0.754 && 0.688 \\
       
&        \UNetVGG~\cite{Mosinska18}  &&  0.636 && 0.328 && 0.607 && 0.776 && 0.748 \\        

&        \UNet{}~ \cite{Ronneberger15} & &
        0.632 &&  % CCQ
        0.323 && % TLTS
        0.619 && % APLS
        0.792 &&  % JUNCT
        0.737 % SUBG
        \\

&
        \DRU~\cite{Wang19c} & &
        0.656 &&  % CCQ
        0.437 && % TLTS
        0.697 && % APLS
        0.821 &&  % JUNCT
        0.768 % SUBG
        \\        
 \cmidrule{2-12}        
&
        \UNetTG{} (Ours) & &
        \bf{0.659} &&  % CCQ
        0.388 && % TLTS
        0.666 && % APLS
        0.808 &&  % JUNCT
        0.767 % SUBG
        \\ 
&
        \DRUTG{} (Ours) & &
        0.657 & &  % CCQ
        0.480 && % TLTS
        \bf{0.725} && % APLS
         \bf{0.837} &&  % JUNCT
        \bf{0.787} % SUBG
        \\

\cmidrule{1-12}

\end{tabular}
}
\caption{
Quantitative comparison between state-of-the-art road network reconstruction algorithms and our proposition. Our proposition \DRUTG{} produces the best results in four out of five metrics while it comes second to \SegP{} in \TLTS{}.
\label{tab:tabel_comp_all}
}
\end{table*}

% !TEX root = ../main.tex
% !TEX spellcheck = en-US

\begin{figure*}[t]
  \centering
  \setlength{\tabcolsep}{3pt}
  \resizebox{1.0\linewidth}{!}{%
\begin{tabular}{cccccc}
  \includegraphics[width=0.155\textwidth]{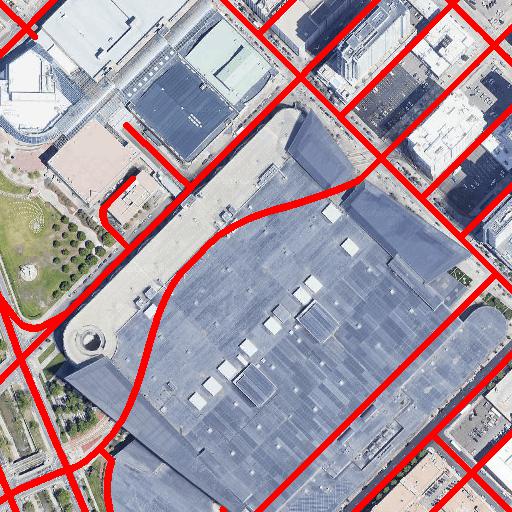} &
    \includegraphics[width=0.155\textwidth]{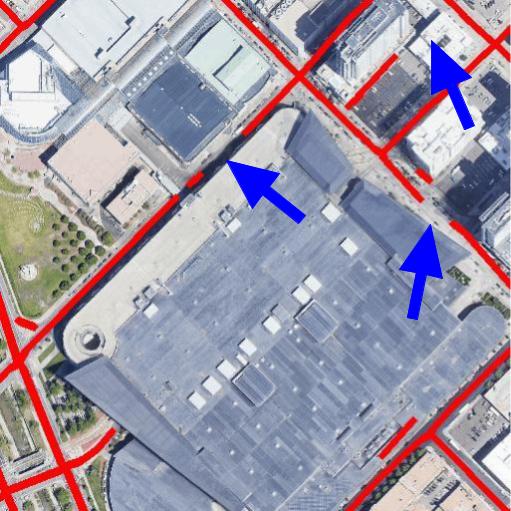} &
  \includegraphics[width=0.155\textwidth]{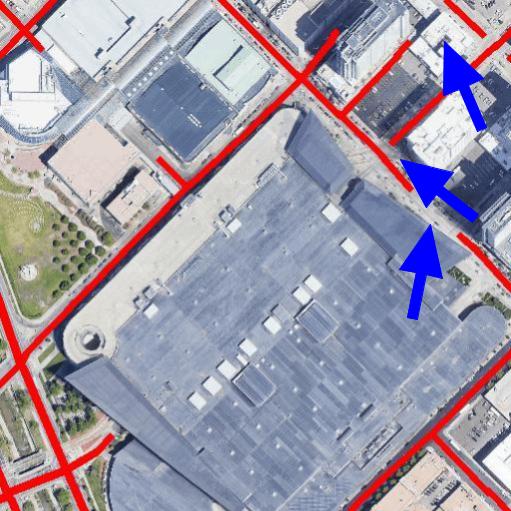}&
  \includegraphics[width=0.155\textwidth]{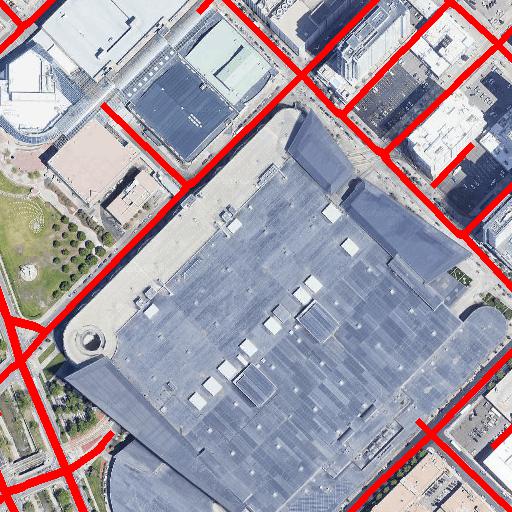} &
    \includegraphics[width=0.155\textwidth]{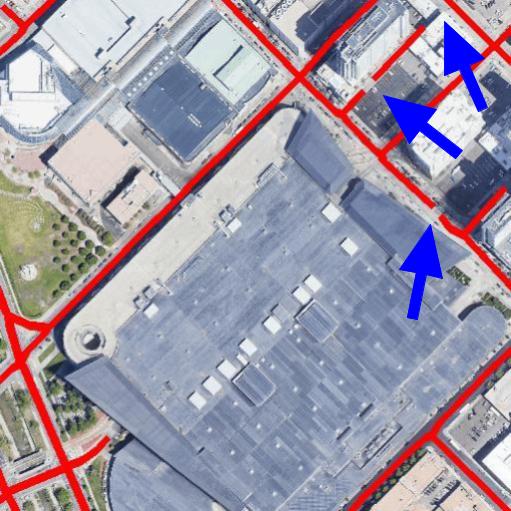}&
 \includegraphics[width=0.155\textwidth]{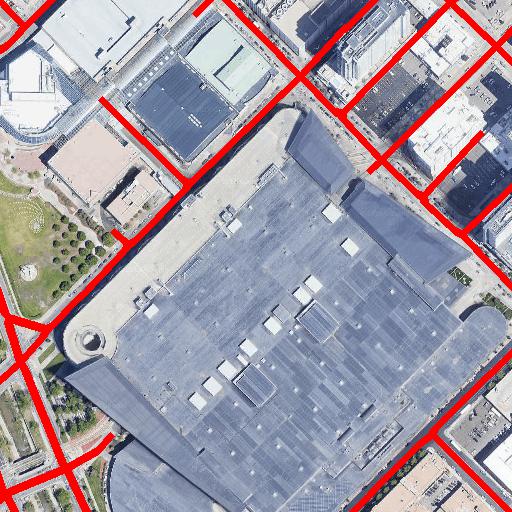}\\
   \scriptsize ground truth& \scriptsize \UNet& \scriptsize {{\it UNet-Van.GAN}}& \scriptsize  \UNetTG& \scriptsize \DRU& \scriptsize \DRUTG\\
   \includegraphics[width=0.155\textwidth]{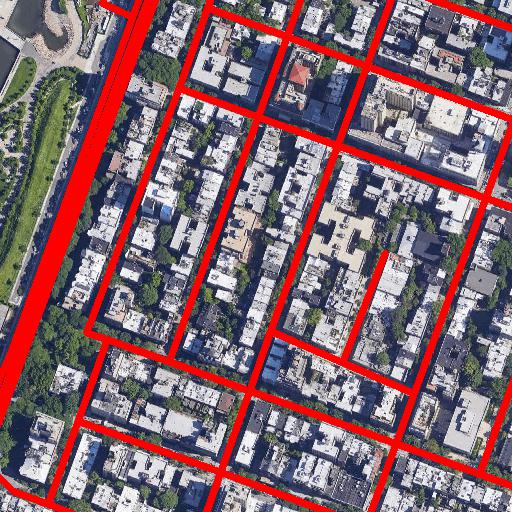} &
  \includegraphics[width=0.155\textwidth]{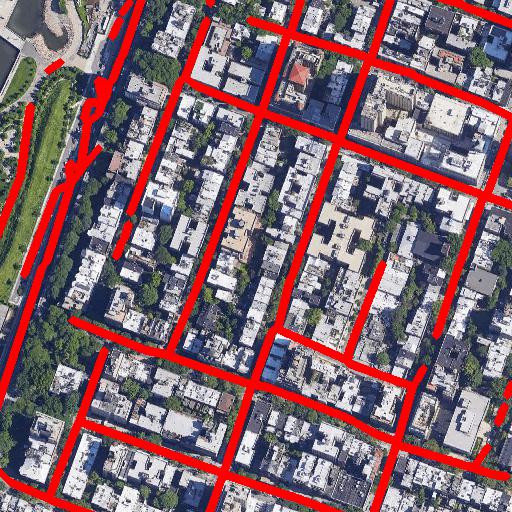} &
  \includegraphics[width=0.155\textwidth]{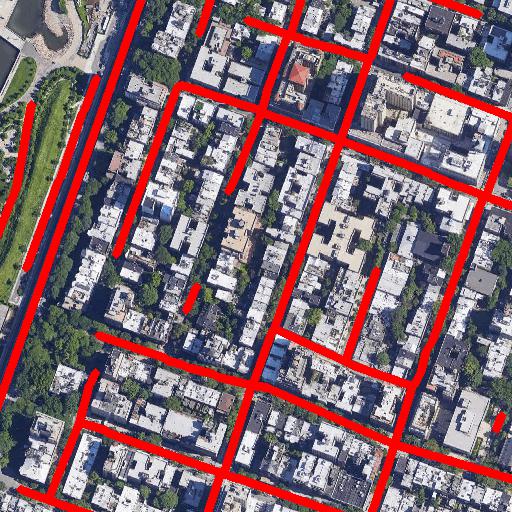} &
   \includegraphics[width=0.155\textwidth]{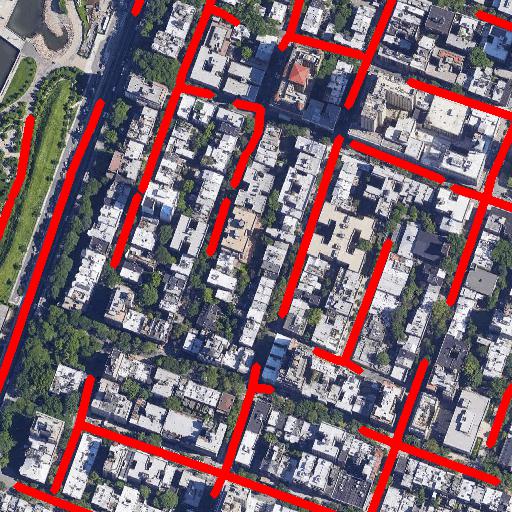}&
    \includegraphics[width=0.155\textwidth]{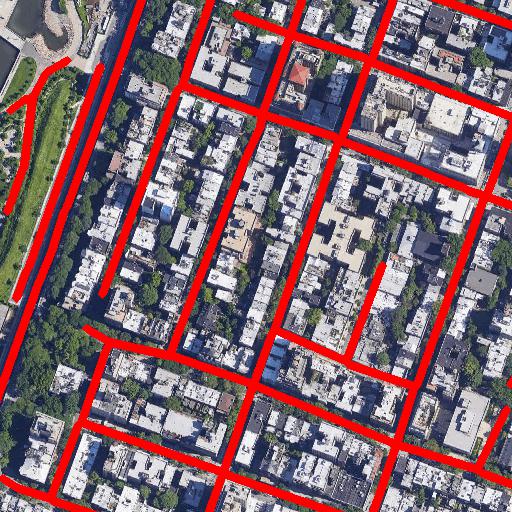}&
   \includegraphics[width=0.155\textwidth]{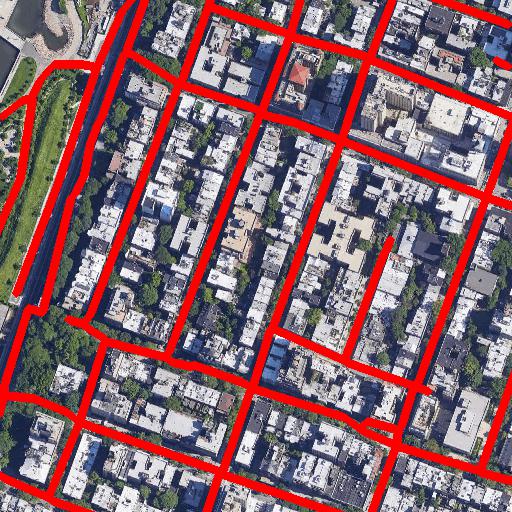}\\
    \scriptsize ground truth& \scriptsize \UNet& \scriptsize \UNetVGG& \scriptsize \RCNN& \scriptsize  \UNetTG& \scriptsize  \DRUTG\\
     \includegraphics[width=0.155\textwidth]{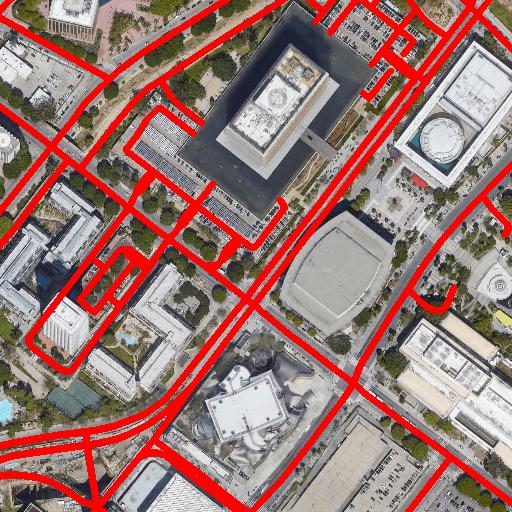} &
  \includegraphics[width=0.155\textwidth]{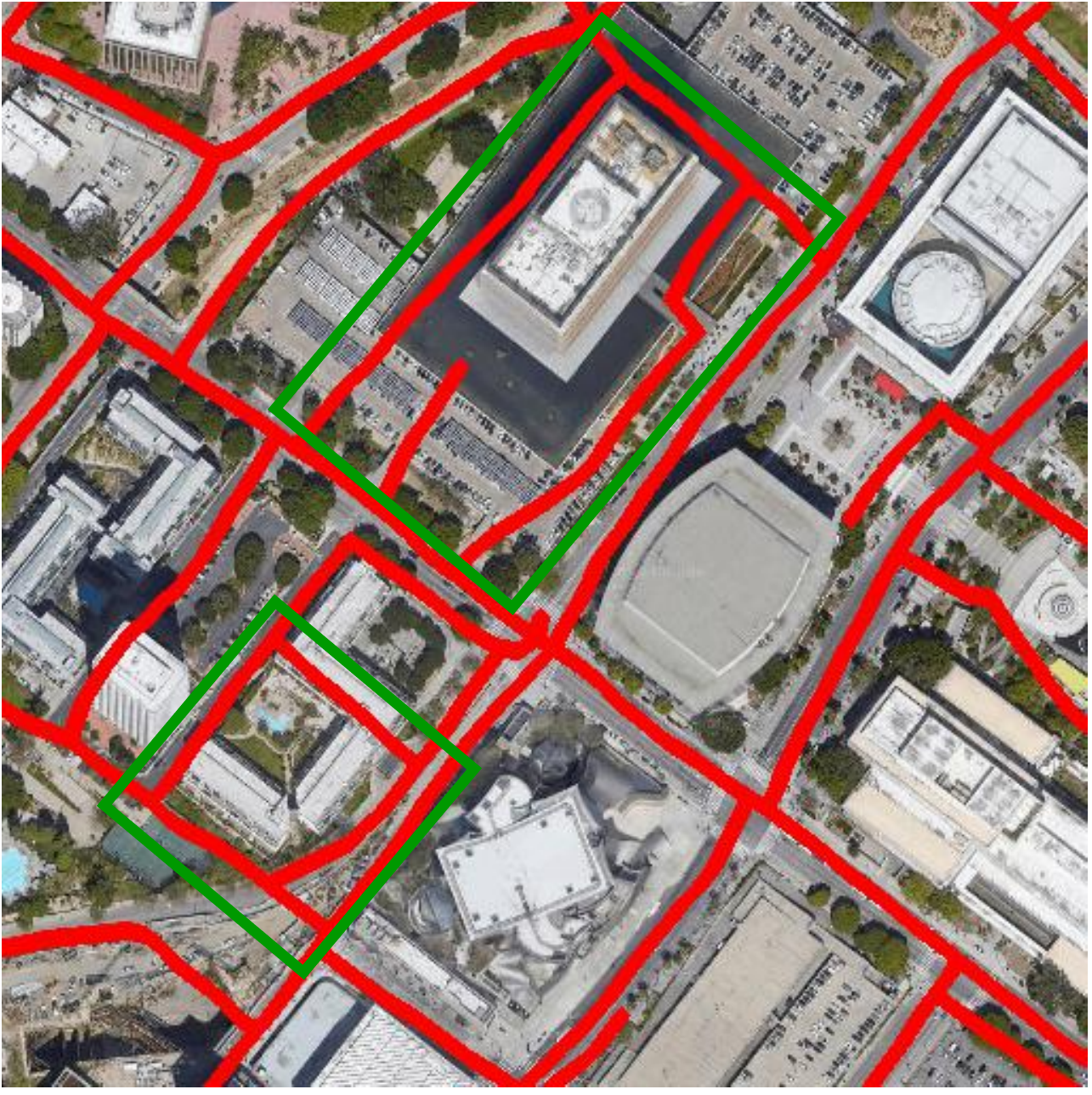} &
  \includegraphics[width=0.155\textwidth]{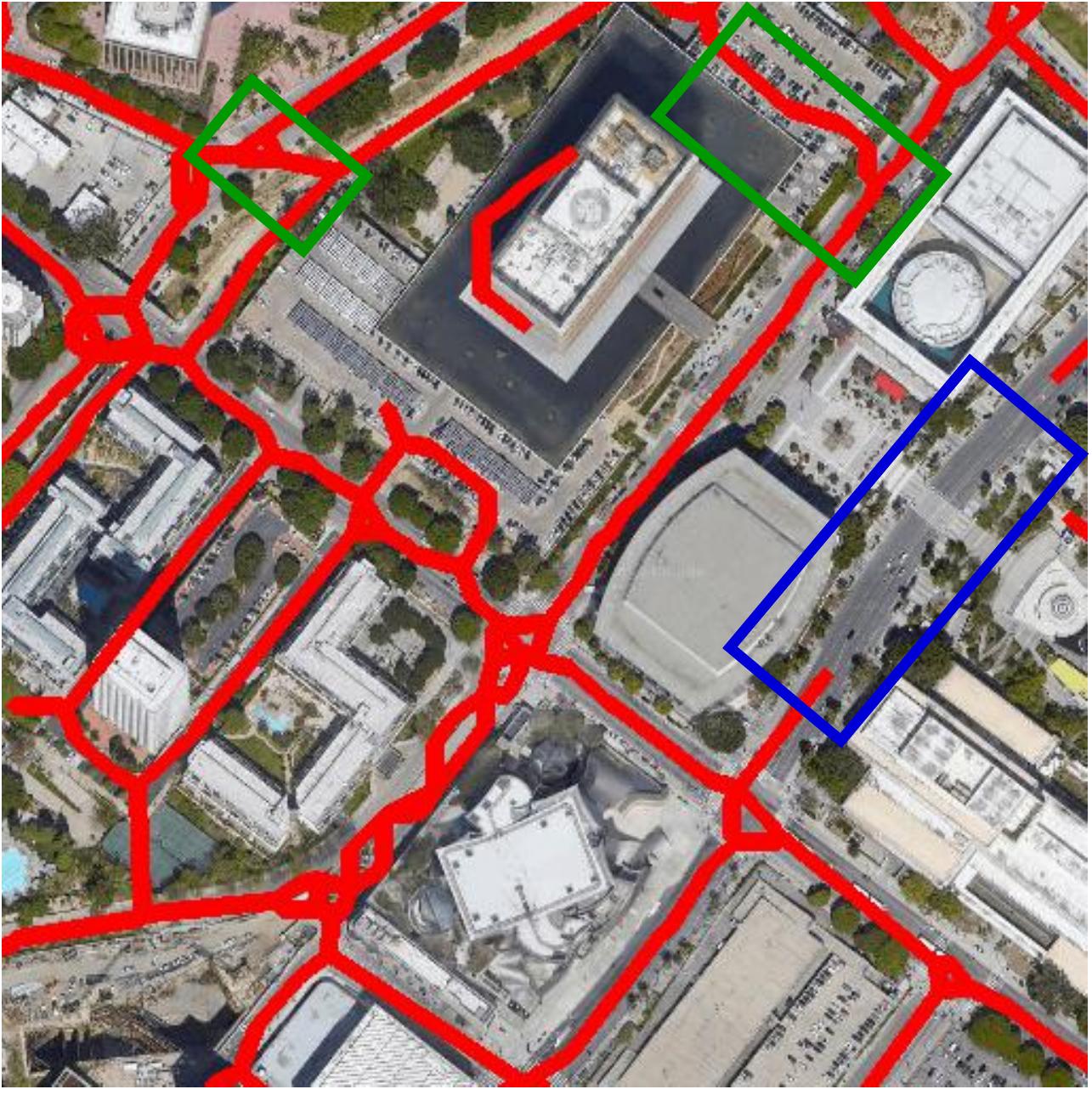} &
   \includegraphics[width=0.155\textwidth]{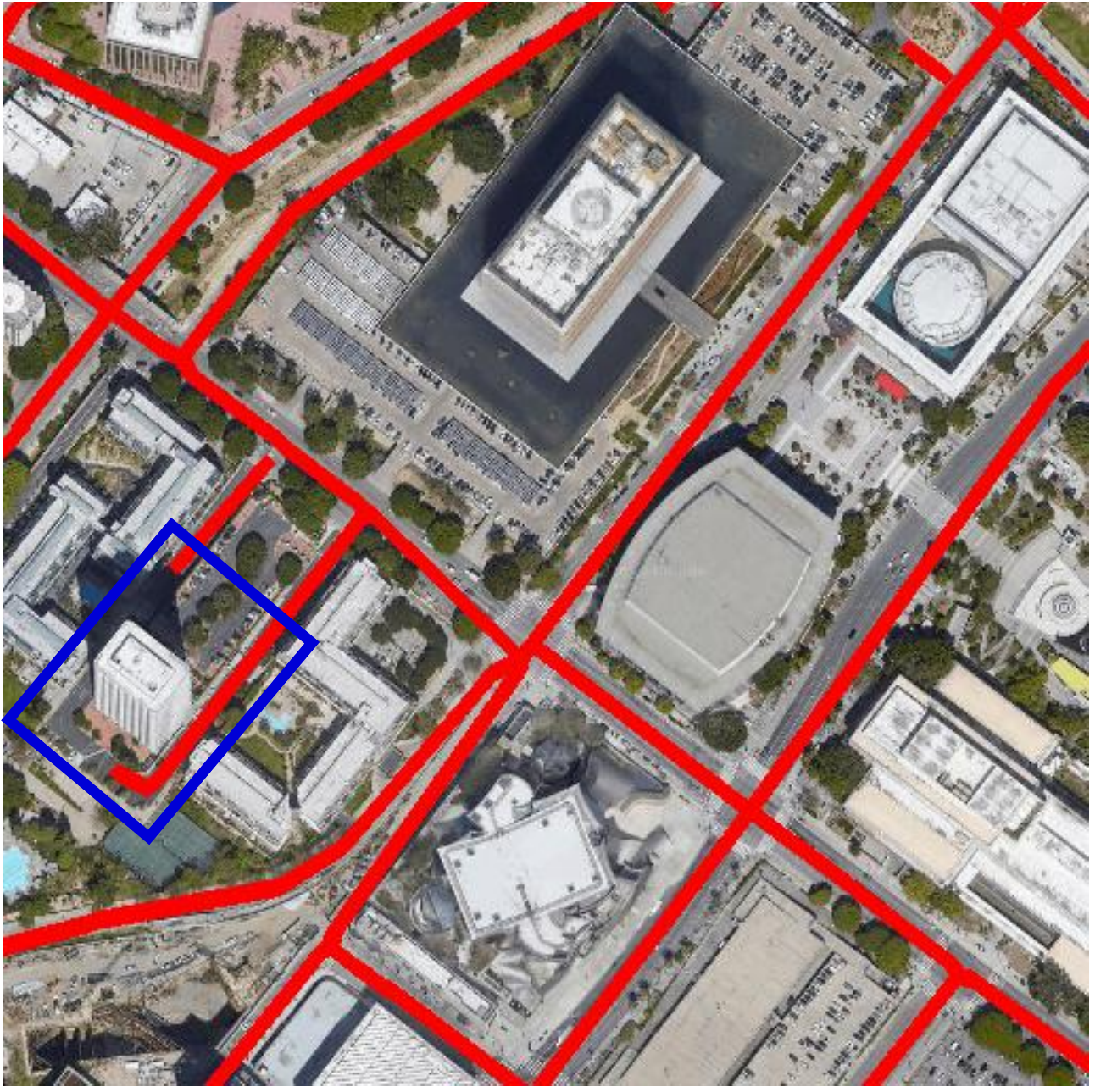}&
    \includegraphics[width=0.155\textwidth]{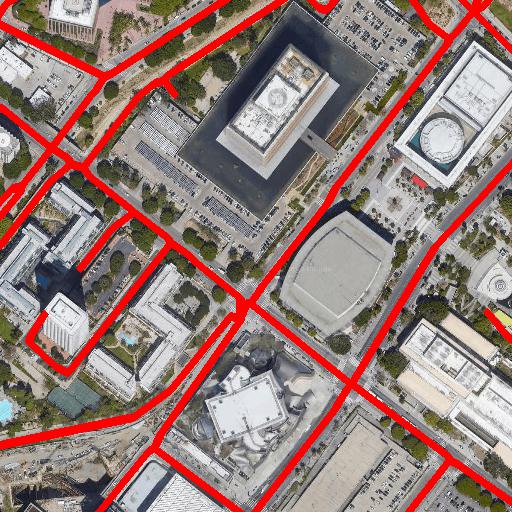}&
   \includegraphics[width=0.155\textwidth]{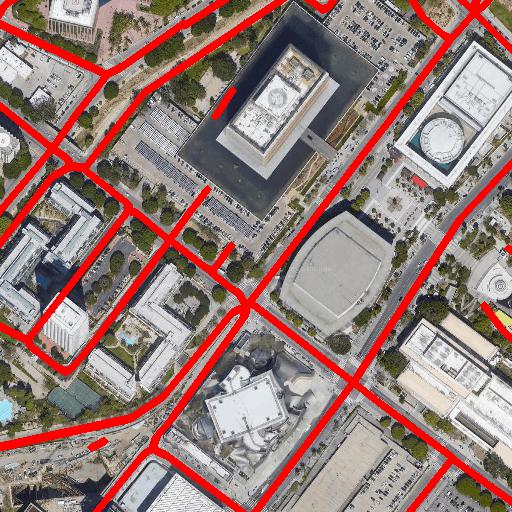}\\
    \scriptsize ground truth& \scriptsize \RT& \scriptsize \SegP& \scriptsize \RC& \scriptsize  \UNetTG& \scriptsize  \DRUTG\\
    \includegraphics[width=0.155\textwidth]{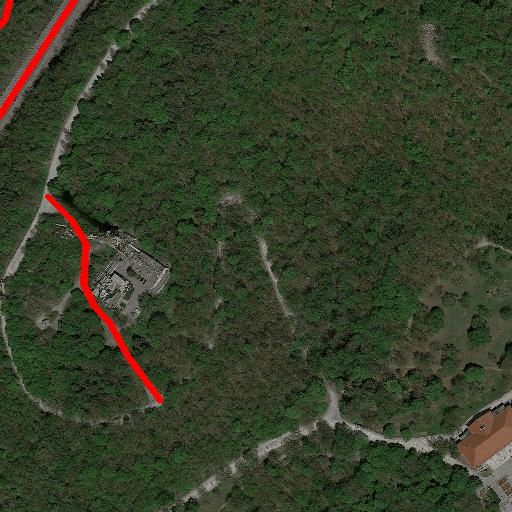} &
  \includegraphics[width=0.155\textwidth]{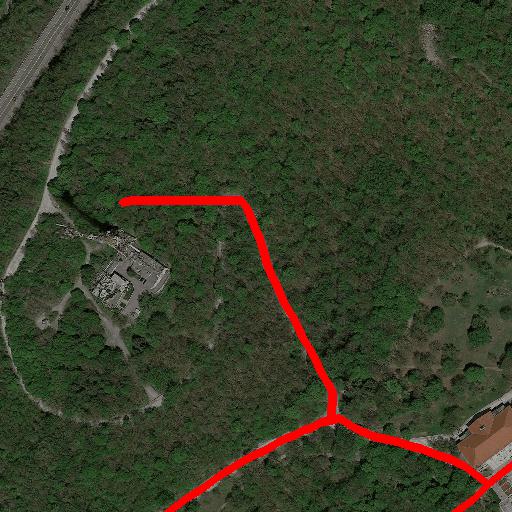} &
  \includegraphics[width=0.155\textwidth]{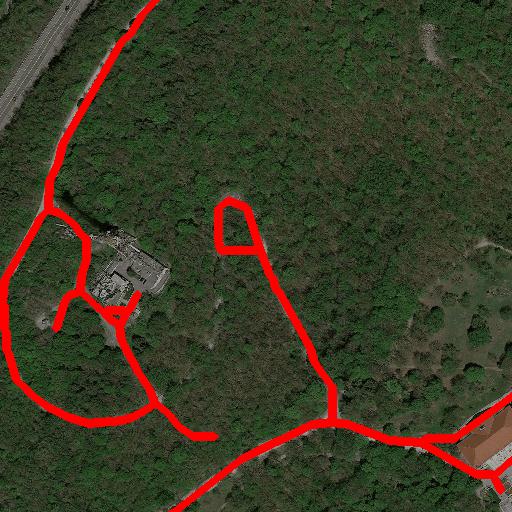} &
   \includegraphics[width=0.155\textwidth]{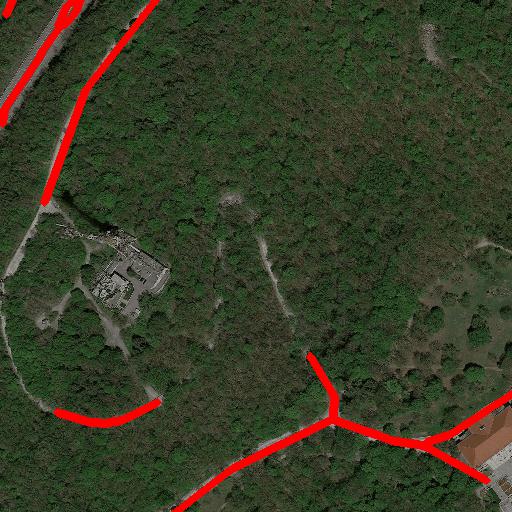}&
  \includegraphics[width=0.155\textwidth]{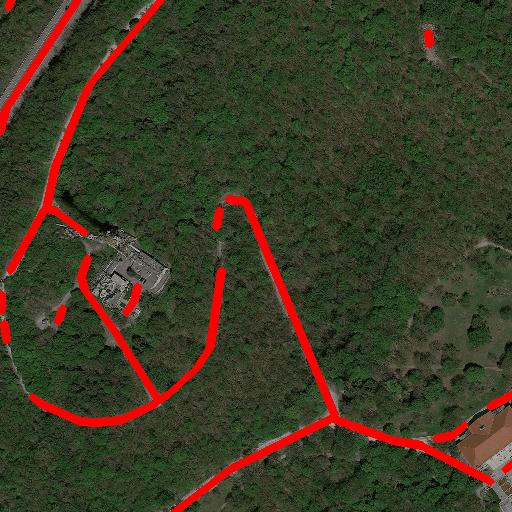}&
 \includegraphics[width=0.155\textwidth]{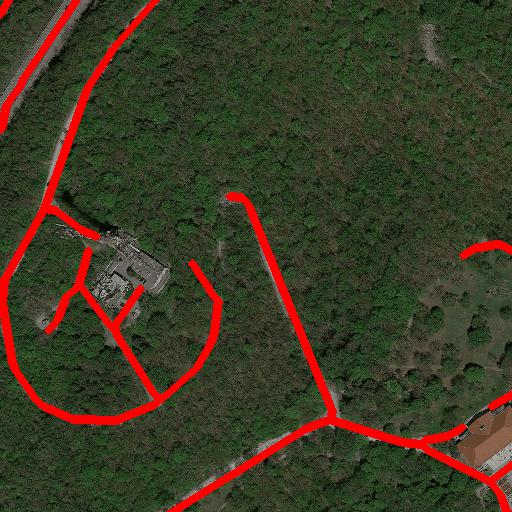}\\
 \scriptsize ground truth& \scriptsize \RT& \scriptsize \SegP& \scriptsize \RC& \scriptsize \DRU& \scriptsize  \DRUTG\\
\end{tabular}
}
  \caption{{\bf Road extraction.} {\small Ground truth and predicted roads are marked in red. Our \TG{} approach improves the topological correctness over their respective generators when used by themselves ($1^{st}$ row), performs better than the segmentation baselines ($2^{nd}$ row) and connectivity-based methods ($3^{rd}$ row), and generalizes well to challenging cases ($4^{th}$ row).}}
  \label{fig:qual_comparison}
\end{figure*}

% !TEX root = ../main.tex
% !TEX spellcheck = en-US

\section{Conclusion}

In this paper, we have proposed an AL strategy that is tailored for extracting networks of curvilinear structures. Its key ingredient is a discriminator that, instead of returning a simple yes or no answer, return a spatially-meaningful descriptor of which parts of the images are well modeled and which are not. The corresponding decisions are made at run-time as opposed to be taken {\it a priori} as in traditional GANs. As a result, we can outperform the state-of-the-art without having to resort to particularly complicated architectures. 

Networks of curvilinear structures---blood vessels, dendrites and axons, bron\-chi, among others---are prevalent in biomedical imagery and recovering their connectivity is also crucial. In future work, we will therefore extend this approach to delineation in 3D image stacks.

\clearpage

\bibliographystyle{splncs04}
\bibliography{5687}
\end{document}